\documentclass{article} 
\usepackage{colm2025_conference} 
\newcommand{\colmfinalcopy}{\colmfinaltrue}\colmfinalcopy 

\usepackage{microtype}
\usepackage{hyperref}
\usepackage{url}
\usepackage{booktabs}

\definecolor{darkblue}{rgb}{0, 0, 0.5}
\hypersetup{colorlinks=true, citecolor=darkblue, linkcolor=darkblue, urlcolor=darkblue}

\title{FREE: The Foundational Semantic Recognition for Modeling \\Environmental Ecosystems}

\author{Shiyuan Luo\\
University of Pittsburgh\\
\texttt{shl298@pitt.edu}
\And
Juntong Ni\\
Emory University\\
\texttt{juntongni02@gmail.com}
\And
Shengyu Chen\\
University of Pittsburgh\\
\texttt{shc160@pitt.edu}
\And
Runlong Yu\\
University of Alabama\\
\texttt{ryu5@ua.edu}
\And
Yiqun Xie\\
University of Maryland\\
\texttt{xie@umd.edu}
\And
Licheng Liu\\
University of Minnesota\\
\texttt{lichengl@umn.edu}
\And
Zhenong Jin\\
University of Minnesota\\
\texttt{jinzn@umn.edu}
\And
Huaxiu Yao\\
University of North Carolina\\
\texttt{huaxiu@cs.unc.edu}
\And
Xiaowei Jia\\
University of Pittsburgh\\
\texttt{xiaowei@pitt.edu}
}

\usepackage{multirow}
\usepackage{booktabs}
\usepackage{booktabs}
\usepackage{colortbl}
\usepackage{amsmath}
\usepackage[table]{xcolor}
\usepackage{graphicx}
\usepackage{wrapfig}

\begin{document}

\ifcolmsubmission
\fi

\maketitle

\begin{abstract}
Modeling environmental ecosystems is critical for the sustainability of our planet, but is extremely challenging due to the complex underlying processes driven by interactions amongst a large number of physical variables. 
As many variables are difficult to measure 
at large scales, existing works often utilize a combination of observable features and locally available measurements or modeled values as input to build models for a specific study region and time period. This raises a fundamental question in advancing the modeling of environmental ecosystems: \textit{how to build a general framework for modeling the complex relationships among diverse environmental variables over space and time? }      
In this paper, we introduce a framework, FREE, that enables the use of varying features and available information to train a universal model. The core idea is to map available environmental data into a text space and then 
convert the traditional predictive modeling task in environmental science to a semantic recognition problem. 
Our evaluation on two societally important real-world applications, stream water temperature prediction and crop yield prediction, demonstrates the superiority of FREE over multiple baselines, even in data-sparse scenarios. 
\end{abstract}

\section{Introduction}

Understanding the dynamics of environmental ecosystems is critical for the sustainable management of natural resources and mitigating natural disasters such as algal blooms and floods, especially given the compelling need for food and water supply from a growing world population and a more unpredictable climate. 
Modeling environmental ecosystems is challenging as these systems are shaped by the complex interactions of a large number of physical variables, such as weather, soil, water, and plants. 
Hence, it often requires the combination of data from the network of weather stations, remote sensing, and field measurements to jointly model complex system dynamics.  
However, many physical variables are often only sparsely available at certain locations or during certain time periods due to the substantial cost required for the data collection.

We focus on two important applications: predicting stream water temperature and annual crop yield under changing weather conditions. Success in both prediction tasks can support a range of applications in optimizing resource allocation and management strategies. 
For example, water reservoir operators in the Delaware River Basin (DRB) need to maintain water temperature within a desired range to ensure safe drinking water for over 15 million people while also preserving sufficiently cool water for aquatic life in downstream areas \citep{williamson2015summary}; the changing climate is making it making it increasingly difficult to sustain high crop yields in the U.S. Corn Belt, threatening food supplies and farmer livelihoods.   
While physics-based models have been developed to simulate underlying physical processes for these ecosystems~\citep{zhou2021quantifying}, the majority of them are necessarily approximations of reality due to incomplete knowledge or excessive complexity in modeling certain processes~\citep{gupta2014debates}. 
Machine learning (ML) models offer an alternative, given their computational efficiency and ability to automatically extract complex data patterns~\citep{he2023physics,rolnick2022tackling,jia2019bringing}.

However, traditional ML approaches face several major challenges in fully leveraging available data for modeling ecosystems: (1) treating physical variables as independent numerical features without explicitly capturing their interdependent physical and ecological relationships. This is further exacerbated by the scarcity of observation data in many real-world ecosystems, which limits the ability of ML models to automatically extract generalizable relationships between features. 
(2) inability to harness inconsistent feature inputs. Current studies on local environmental ecosystems often enhance the prediction of their target study region by leveraging multiple types of input features related to the target variable, and the selected features often vary between different studies. Besides meteorological data (e.g., solar radiation) that are commonly used, prior works also explored including other measurements on the environment system (e.g., soil properties~\citep{chen2023physics}, land use and geometric structures~\citep{cho2022improving}), and other physical variables simulated by physics-based models~\citep{jia2021physics_sdm}. These features may not always be available for data samples collected from different locations and time periods, which poses a challenge in training a global model that can utilize different input features. (3)~lack of a general pipeline that can dynamically incorporate auxiliary observations into the ML model. 
Current works on assimilating auxiliary observations rely on task-specific learning mechanisms or model structures, e.g., Kalman filtering for incorporating new observations~\citep{zwart2023near}, and graph convolution and invertible network layers for assimilating observations from neighboring samples~\citep{chen2021heterogeneous, brajard2020combining}. Yet these approaches can be computationally expensive to implement when we have a large state space and/or non-linear system dynamics.
Recent LLM advances show promise for addressing these limitations~\citep{mai2023opportunities}. Originally developed for language tasks, LLMs now handle tabular data by converting it to text~\citep{wang2023anypredict, zhao2023large}, enabling flexible feature handling and robustness to missing values.

In this paper, we propose a novel method, \textbf{F}oundational semantic \textbf{R}ecognition for modeling \textbf{E}nvironmental \textbf{E}cosystems (FREE) to address these limitations.
The key idea of FREE is to translate the heterogeneous input data into natural language descriptions using large language models and then estimate the target variable through semantic recognition in the text space. This addresses the challenges in utilizing different data sources by only manipulating the text space while maintaining the same predictive modeling component (i.e., semantic recognition). 
In particular, the translation process uses the available features for each data point, allowing the set of features to vary across data points. Moreover, the translation process can easily incorporate auxiliary observations (e.g., newly collected observations from the previous day) into the textual description with a properly designed prompt. 
After obtaining the textual description, we build the semantic recognition component by leveraging a separate language model to embed the text data and incorporating additional network layers (e.g., long-short term memory (LSTM)) to capture temporal dependencies. The use of the language model on textual descriptions enables a better understanding of the nature and semantics of input features. To further enhance the embedding performance of the language model (LM) on environmental descriptions, we pre-train the semantic recognition component using abundant simulated samples generated by physics-based models. 
This pre-training process also helps the model learn the general physical relationships encoded in the physics-based models and mitigate the challenge posed by the sparse observations.

We evaluate FREE on two real-world tasks: daily stream water temperature prediction in the DRB, and annual county-wise corn yield prediction in Illinois and Iowa. Both of them cover a diverse set of locations and long time periods. 
FREE shows its superior predictive performance over multiple baselines, especially with sparse observations, and it can effectively handle different input features and incorporate auxiliary observations. The pre-training process also helps improve the prediction 
performance of the model when adapted to different locations. 
This work also opens new opportunities to leverage LLMs as a universal tool for general environmental modeling tasks, which are currently often addressed in a fragmented and localized manner due to variations in data collection and processing pipelines. 
\section{Method}


\begin{figure*}
    \centering
    \includegraphics[width=0.9\linewidth]{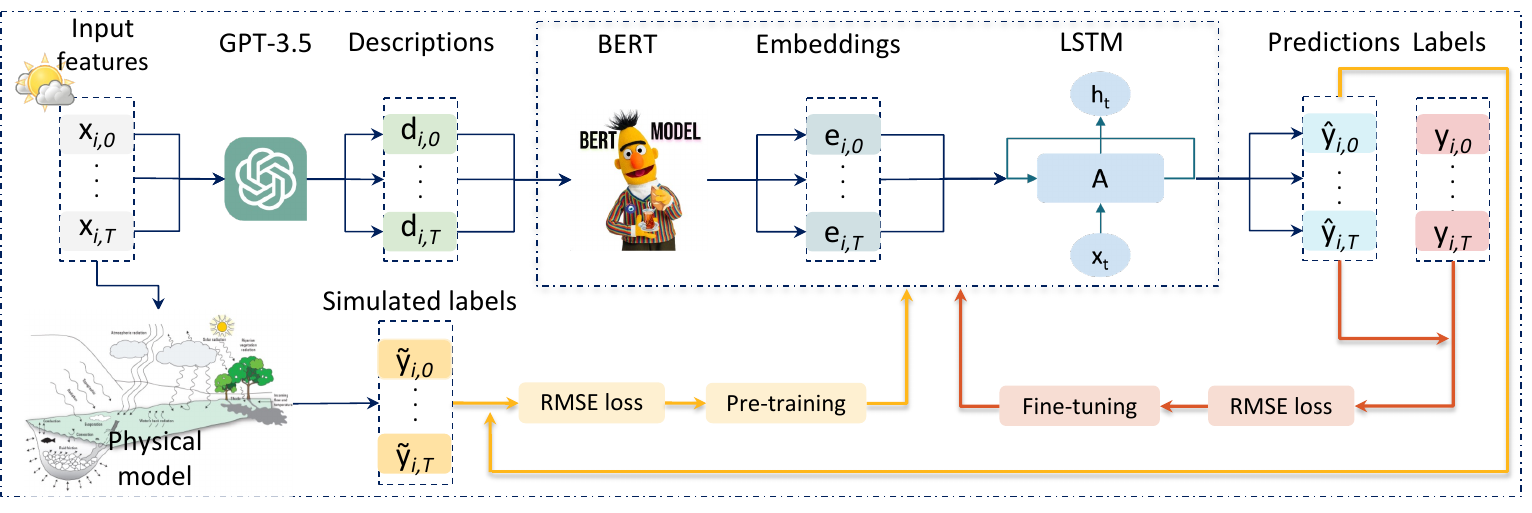}
    \caption{The framework of FREE: Input features are first transformed into natural language descriptions by LLMs. These descriptions are then processed by a LM to generate embeddings, which are fed to an LSTM layer for making predictions. Simulated labels generated by a physics-based model are used to pre-train the LM and LSTM layers, followed by fine-tuning with true observations of the target variable for enhanced predictions.}
    \label{fig:flow_chart}
\end{figure*}

In this section, we define the problem and introduce our proposed framework. 
The objective is to predict the target variable $y$ (e.g., water temperature or crop yield) at multiple locations $i \in \{1,..., N\}$ and over a period of time. 
For clarity, we denote by $\mathbf{x}_{i,t}$ the input features at the location $i$ on date $t$. 
The input features contain the set of meteorological variables (e.g., solar radiation, rainfall) that are commonly used as drivers for physics-based models. 
Besides, we include some other physical variables estimated by physics-based models (e.g., cloud cover fraction, groundwater properties) or obtained from other data sources (e.g., soil properties).  
Notably, these features may be absent for certain locations or time steps. 
We represent the observations of the target variable at each location $i$ over multiple time steps as $y_i = \{{y}_{i,1}, {y}_{i,2}, ..., {y}_{i,T}\}$. The observations can be only sparsely available for certain time steps and locations. 

FREE converts the traditional predictive modeling task in environmental science to the semantic recognition problem in a text space created by LLMs (e.g., GPT). 
A clear benefit of this framework is its ability to harness inputs of different feature sets and incorporate auxiliary information. We also introduce additional model components to embed the text data and capture the temporal dependencies. To mitigate the need for large training samples 
for different ecosystems, we pre-train the model using abundant simulated data generated by physics-based models. In the following, we will describe these components in detail. 




\subsection{Overall architecture}
The FREE framework (shown in Fig.~\ref{fig:flow_chart}) 
consists of two major components in its architecture. 


\textbf{Input data conversion: }
To address inconsistencies in the feature set and incorporate auxiliary observations, we propose to transform the original data sample into a corresponding natural language description. 
This approach facilitates the handling of diverse and potentially incomplete feature sets for different data points, enabling a uniform textual representation of data across varying input scenarios.
Specifically, we leverage an existing LLM to transform each data point $\mathbf{x}_{i,t}$ into clear, natural language descriptions $d_{i,t}$. 
To effectively communicate with the LLM, we use a linearization technique~\citep{wang2023anypredict} to construct prompts that consist of a context-setting prefix, the linearized data input, and a directive suffix. The prefix ($p$) provides the model with a background of the dataset, while the suffix ($s$) instructs the model on how to format its output. The complete prompt is formulated as:
\begin{equation}
\small
d_{i,t} = \text{LLM}\left(p, \text{linearize}(\mathbf{x}_{i,t}), s\right). 
\end{equation}

\begin{wrapfigure}[20]{r}{0.6\textwidth}
    \centering
    \includegraphics[width=0.6\textwidth]{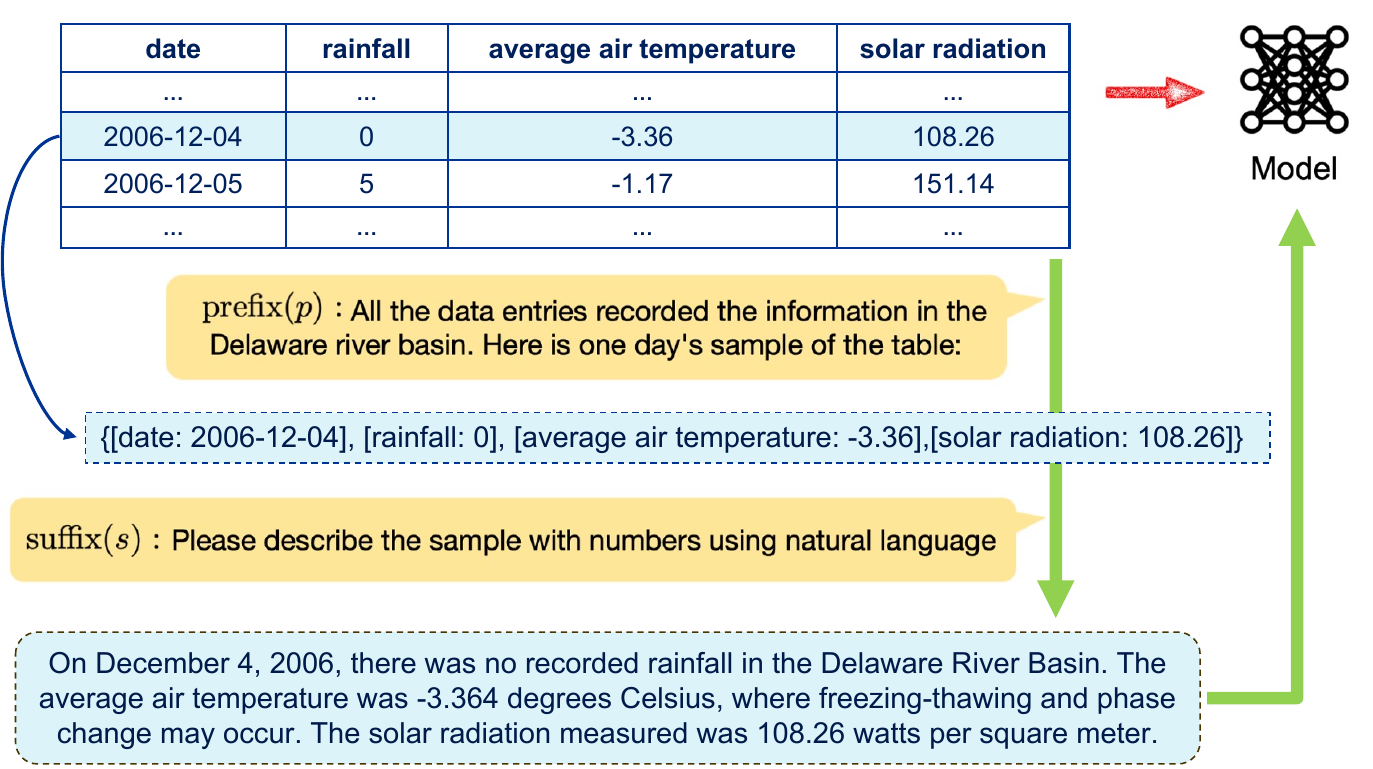}
    \caption{Green arrows indicate FREE handling inputs of diverse feature sets that use linearized data. Red arrow suggests that traditional ML models might need separate preprocessing methodologies to address data irregularities effectively. }
    \label{fig:conversion}
\end{wrapfigure}

As depicted in Fig.~\ref{fig:conversion},  we format the input features $\mathbf{x}_{i,t}$  of $K$ dimensions/variables into a sequence pairing column names ${c}^k_{i,t}$ with the corresponding feature values $\mathbf{x}^k_{i,t}$ (e.g., \{\,[rainfall: 0], [solar radiation: 151.14]\,\}\,). This process can be expressed as follows:  
\begin{equation}
\small
\text{linearize}(\mathbf{x}_{i,t})= \{[{c}^k_{i,t}:\mathbf{x}^k_{i,t}]\}_{k=1}^K.
\end{equation}

Given the prompt, LLM can generate feature summaries that are both descriptive and succinct.
, such as: "\textit{On December 4, 2006, there was no recorded rainfall in the Delaware River Basin. The average air temperature was -3.36 degrees Celsius, where freezing-thawing and phase change may occur. The solar radiation measured was 108.26 watts per square meter.}". 
The obtained textual descriptions, alongside the observations in the original dataset, form paired data samples 
as $\{d_{i,t}, y_{i,t}\}$, which are then used for tuning the semantic recognition component (to be discussed later). 

In summary, the proposed input translation process helps generate an understandable and flexible representation of input features. More importantly, the use of LLM enables supplementing the raw feature values with semantically meaningful text descriptions and interpreting underlying physical phenomena, which facilitates subsequent modeling components to capture the complex nature and the interactions of input features.


\textbf{Semantic recognition:}
Once obtaining the textual descriptions $d_{i,t}$, we utilize a separate LM $f$ to process the obtained text data and embed them as 
\begin{equation}
\small
\mathbf{e}_{i,t}=f(d_{i,t}). 
\label{eq:embed}
\end{equation}

In this work, we use the DistilBERT model~\citep{sanh2019distilbert} in this step. 
It is vital to note that the embeddings $\mathbf{e}_{i,t}$ are generated independently across different data points. 
To effectively capture the intricate data temporal dependencies, we introduce additional network layers that combine multiple data embeddings. In particular, this work uses LSTM layers to capture common temporal data dependencies in the environmental ecosystem, e.g., seasonal changes, and the effect of heavy rainfall on soil water in the next few days. 



\subsection{Handling diverse inputs and auxiliary data}
\label{subsec: incorporate}

One major goal of modeling environmental ecosystems is long-term prediction in large regions. FREE provides opportunities to enhance ML models towards this goal from two aspects.  
First, predictive models often have larger predictive errors as time progresses in the test phase due to the accumulated bias in the model state. 
One could explore leveraging newly collected observations during the testing phase (e.g., the observations collected at previous time $t-1$) to mitigate the current model bias and improve the future predictions (e.g., the prediction after $t$). 
Second, 
when training the model across space, some areas may offer more features, such as certain variables related to local soil properties while other regions may only provide weather-related features. 
Such discrepancies in feature availability pose challenges for building a universal model across space. 
Under the FREE framework, we can leverage additional observations and handle different features through simple modification of the prompt while keeping the predictive model (i.e., the semantic recognition component) unchanged.

\textbf{Incorporating auxiliary observations:}
In this study, we consider incorporating two types of auxiliary observations to enhance the prediction of the data sample $\mathbf{x}_{i,t}$: the newly collected observation from the previous time $t-1$ and at the same location $i$, and from the neighboring locations of the target location $i$.  
We incorporate the auxiliary data by updating the linearized data input to include it.
As observations are collected from different dates with the existing features in $\mathbf{x}_{i,t}$, we need to explicitly include the exact date information for the auxiliary observations and original features. 
If we consider the auxiliary observations being collected at the date $t-1$ from both the current location $i$ and its neighbor $j\in \mathcal{N}(i)$, the updated linearization process can be expressed as:
\begin{equation}
\small
\begin{aligned}
&\text{linearize}([\mathbf{x}_{i,t},y_{i,t-1},y_{j,t-1}]_{j\in \mathcal{N}(i)})\\
= &  [\text{date}:t-1]\cup[{c}_i^y:y_{i,t-1}]\cup[{c}^y_j:y_{j,t-1}] \cup [\text{date}:t] \cup\{[{c}^k_{i,t}:\mathbf{x}^k_{i,t}]\}_{k=1}^K,
\end{aligned}
\end{equation}
where $c_i^y$ denotes the column name for the observed labels, $c_j^y$ denotes the column name for the observed labels from the neighboring location $j$, and $\cup$ indicates concatenation operation across the sequences.
LLM then follows the instructions by $p$ and $s$ to generate the description using these data. 
It is noteworthy that the auxiliary observations may not always be available.
When creating the linearized data, we include the columns of the auxiliary observations
only if they are available, and skip them otherwise. 

\textbf{Handling different input features:}
FREE handles different input features by linearizing only available features, skipping missing ones. 
In particular, if a feature $k$ is not available for the data sample $\mathbf{x}_{i,t}$, then the pair $[c^k_{i,t}: \mathbf{x}^k_{i,t}]$  will be skipped and not included in the input to the LLM. 
This enables the use of a combination of heterogeneous data samples with different feature sets in both training and testing processes. 
This also allows the subsequent semantic recognition to proceed seamlessly on the generated text without the need for manual adjustments to account for data irregularities.

\subsection{Pre-training using physical simulations}

\begin{wrapfigure}[12]{r}{0.5\textwidth}
    \centering
    \includegraphics[width=0.5\textwidth]{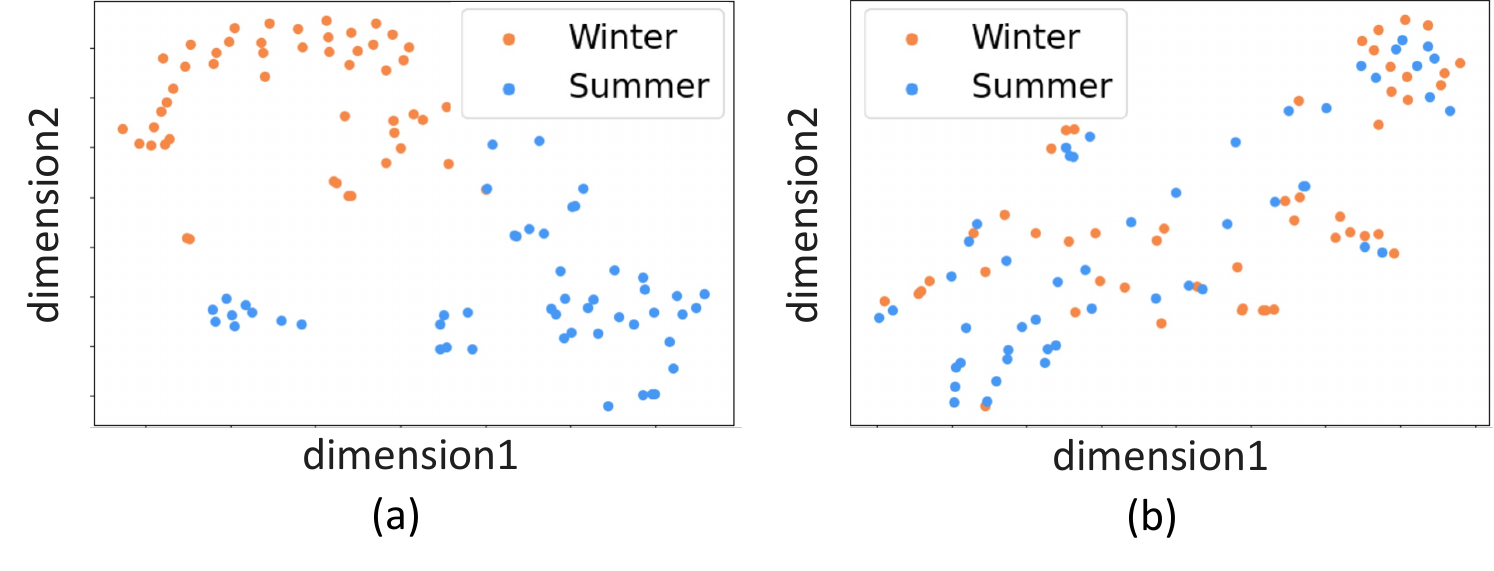}
    \caption{t-SNE of embeddings for data points randomly sampled from summer and winter.}
    \label{fig:cluster}
\end{wrapfigure}
The LM in the semantic recognition phase (Eq.~\ref{eq:embed}) are not inherently trained on data specific to the target environmental ecosystems.
As shown in Fig.~\ref{fig:cluster}~(b), the input features samples from different seasons tend to be mixed in the latent embedding space of the original LM, when applied directly, may fall short of capturing semantics from the text generated for our target task. 

Tuning the LM towards the target domain requires sufficient observed samples, which are often not available in real-world ecosystems. 
Therefore, we use simulated data $\tilde{y}_{i,t}$ generated by physics-based model to pre-train the semantic recognition component. 
Fig.~\ref{fig:cluster} (a) shows that samples of different seasons can now be distinguished, confirming that the LM model tuned with simulated data can better capture semantics in the textual descriptions $d_{i,t}$. 

This simulation-based pre-training also facilitates the training of the semantic recognition component to emulate general physical processes encoded in the physics-based model, enhancing the model's generalizability as many physical processes generally hold across space and time. 
Upon completing this pre-training, it requires only a few epochs of fine-tuning using true observations 
before reaching a quality model.  

\section{Experiments}

\begin{table}[!t]
\small
\centering
\caption{Prediction RMSE for stream temperature using 1\%, 2\%, 4\%, and 100\% randomly selected training labels. The best results are \textbf{bold}, the second best results are \underline{underlined}.}
\begin{tabular}{c|l|c|c|c|c}
\toprule
\textbf{Dataset} & \textbf{Method} & \textbf{100\%} & \textbf{4\%} & \textbf{2\%} & \textbf{1\%}\\ 
\midrule
\multirow{4}{*}{SS} & LSTM & 1.90 & \underline{1.95} & \underline{2.14} & \underline{2.23} \\ 
                    & Transformer & \underline{1.79} & 2.05 & 2.20 & 2.27 \\
\cmidrule{2-6} 
                  & FREE & \textbf{1.70} & \textbf{1.71} & \textbf{1.90} & \textbf{2.09} \\
\midrule
\multirow{8}{*}{CRW} & LSTM & 1.90 & 2.33 & 2.80 & 2.87 \\
                    & Transformer & 2.14 & \underline{2.20} & \underline{2.19} & \underline{2.30} \\
                  & RGRN & \underline{1.78} & 2.31 & 2.61 & 2.80 \\
                  & Gr-CNN & 1.80 & 2.38 & 2.69 & 3.56 \\
                  & HydroNets & 1.87 & 2.51 & 2.77 & 3.75 \\
\cmidrule{2-6} 
                  & FREE & \textbf{1.61} & \textbf{1.68} & \textbf{1.62} & \textbf{1.65} \\
\bottomrule
\end{tabular}
\label{fig:delware1}
\end{table}

\begin{table}[!t]
\small
\centering
\caption{Prediction RMSE for annual corn yield using 10\%, 20\%, 50\%, and 100\% randomly selected training labels. The best results are \textbf{bold}, while the second best results are \underline{underlined}.}
\begin{tabular}{c|l|c|c|c|c}
\toprule
\textbf{Dataset} & \textbf{Method} & \textbf{100\%} & \textbf{50\%} & \textbf{20\%} & \textbf{10\%}\\ 
\midrule
\multirow{4}{*}{Crop} & LSTM & \underline{73.56} & 79.67 & 78.88 & 84.52 \\ 
                      & PG-GNN & 79.92 & \underline{71.68} & \underline{72.72} & \underline{79.04} \\
                      & PG-AN & 96.83 & 114.40 & 90.52  & 100.36 \\
                      & Transformer & 84.93 & 79.78 & 97.56 & 101.40\\

\cmidrule{2-6} 
                  & FREE & \textbf{65.24} & \textbf{63.74} & \textbf{63.75} & \textbf{64.28} \\
\bottomrule

\end{tabular}
\label{fig:crop}
\end{table}

In this section, we present our datasets and provide a comprehensive assessment of the proposed methods. Our primary emphasis lies on predictive performance using sparse data, underscoring the efficacy of the proposed FREE framework and the simulation-based pre-training. Besides, we delve into additional experiments that validate our model's capability to harness diverse input features and auxiliary observations. 
We further examine its performance under transfer learning scenarios.

\subsection{Experimental Setup}

\label{sec:DatasetsAndBaselines}
\paragraph{Stream dataset and baselines}
We use data from the Delaware River Basin (DRB), an ecologically diverse region and a watershed along the east coast of US. We study two subsets: Christina River Watershed (\textbf{CRW}) (with 42 connected river segments), and a single stream segment from a distinct region (\textbf{SS}). 
We train each ML model using data from October 31, 2006 to July 8, 2013 (2,450 days), and test on the following 2,450 days, up to March 30, 2020. 
On a daily scale, we incorporated basic meteorological features (the day of the year, rainfall, daily average air temperature, and solar radiation). In addition to these, we also considered other features, namely average cloud cover fraction, groundwater temperature, subsurface temperature, and potential evapotranspiration.
We compare FREE with multiple popular baselines developed for stream modeling, including LSTM model (considered to be the state of the art in many environmental modeling problems~\citep{shen2021applications}), Transformer~\citep{vaswani2017attention}, RGRN~\citep{jia2021physics_sdm},Gr-CNN~\citep{sun2021explore}, HydroNets~\citep{moshe2020hydronets}. Besides, we include a baseline that assimilates new observations for updating the model, HRGN-DA~\citep{chen2021heterogeneous}.

\paragraph{Crop dataset and baselines} 
We use the corn yield data in Illinois and Iowa from the years 2000-2020 provided by USDA National Agricultural Statistics Service (NASS) (available at \url{https://quickstats.nass.usda.gov/}). 
The observed annual crop yield labels cover 199 countries spanning 21 years from 2000-2020, with daily records for each year. 
The input data include seven features: the surface downward shortwave radiation, air temperature, humidity, wind speed, precipitation, depth-weighted averaged bulk density in the soil, and depth-weighted averaged sand content in the soil. 
To reduce computational load, we randomly sample 1/7 of synthetic data in the pre-training phase. We use the observational data from 2000-2017 for training and the data from 2018-2020 for testing.
We compare the proposed method with the LSTM and other two baselines that have both proven effective in crop yield prediction, i.e.,  PG-AN~\citep{he2023physics}, and PG-GNN~\citep{fan2022gnn}.

\subsection{Overall predictive performance}
\begin{wrapfigure}[22]{r}{0.5\textwidth}
    \centering
    \includegraphics[width=0.5\textwidth]{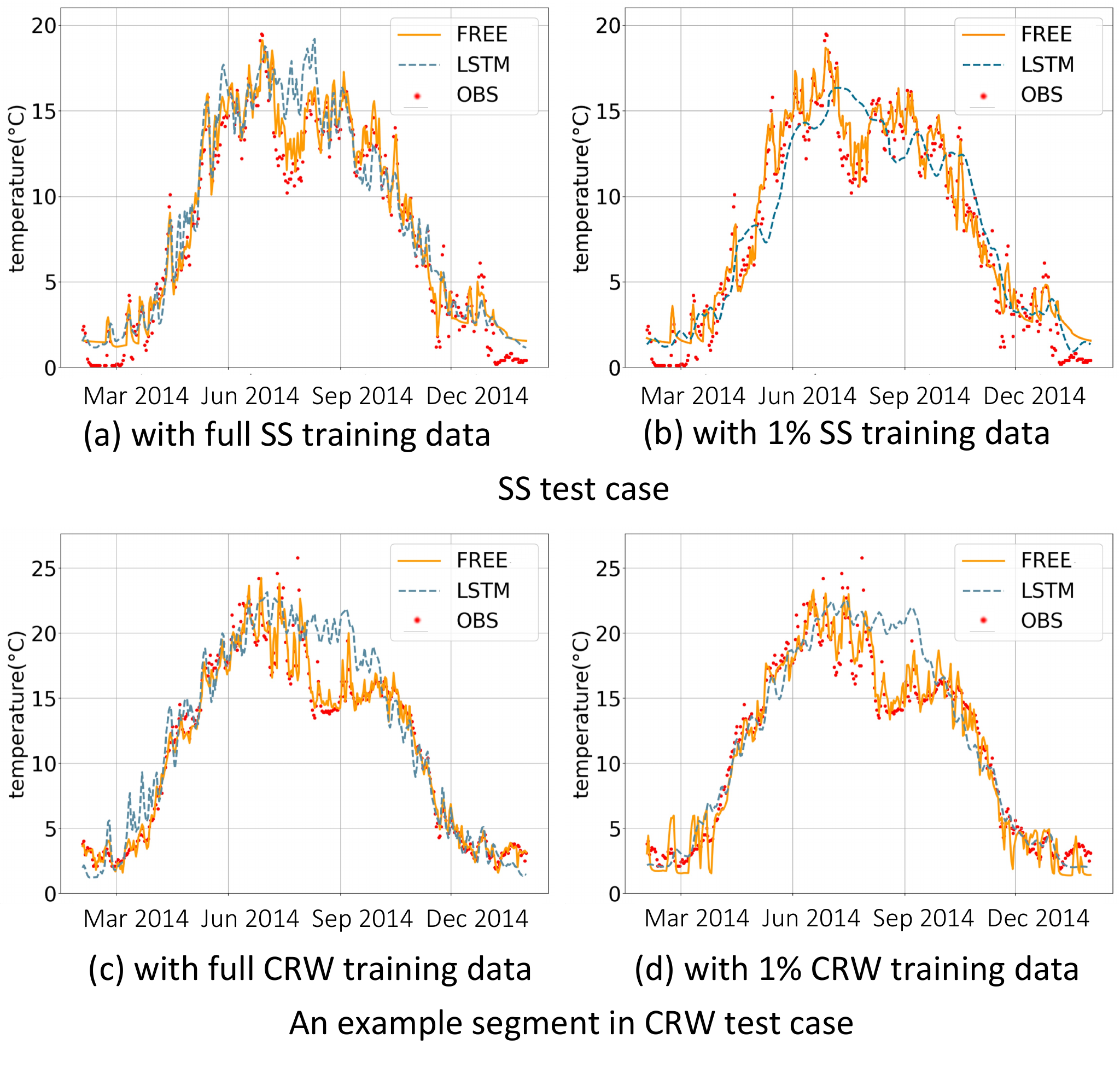}
    \caption{\small Comparison of FREE and LSTM on stream water temperature prediction.}
    \label{fig:pred_obs}
\end{wrapfigure}
Tables~\ref{fig:delware1} and~\ref{fig:crop} show the performance of various methods for predicting stream water temperature and annual crop yield, respectively. The notation 
FREE represents a model pre-trained on synthetic data and then fine-tuned on observational data. 
Table~\ref{fig:delware1} shows that FREE outperforms all baselines when data are abundant, showing its effectiveness in capturing the semantics and relationships of input features from textual descriptions. 
Notably, the FREE maintains good performance regardless of label volume, while other models generally have much worse performance in data-sparse scenarios. This underscores the model enhancement brought by the simulation-based pre-training in learning generalizable and domain-specific semantics. 
Moreover, pre-training our model 
significantly reduces the time for effective fine-tuning (about 
one-tenth of the time compared to direct fine-tuning). 
Fig.~\ref{fig:pred_obs} shows the predicted water temperature over one year for the SS and a river segment within the CRW to demonstrate the alignment of predictions and true observations. 
The figure clearly shows that our method outperforms the baseline model under different data sparsity scenarios (100\% and 1\%), especially in capturing the water temperature oscillations in the summertime.
The LSTM's predictions, while reasonably tracking the general trend, fall short in replicating the nuanced variations, especially as the training data becomes sparser. LSTM struggles to recognize subtle changes and creates smooth predictions with limited data, overlooking smaller peaks and valleys that our method consistently identifies and traces. 

\subsection{Handling auxiliary information}
\begin{wrapfigure}[17]{r}{0.5\textwidth}
    \centering
    \includegraphics[width=0.5\textwidth]{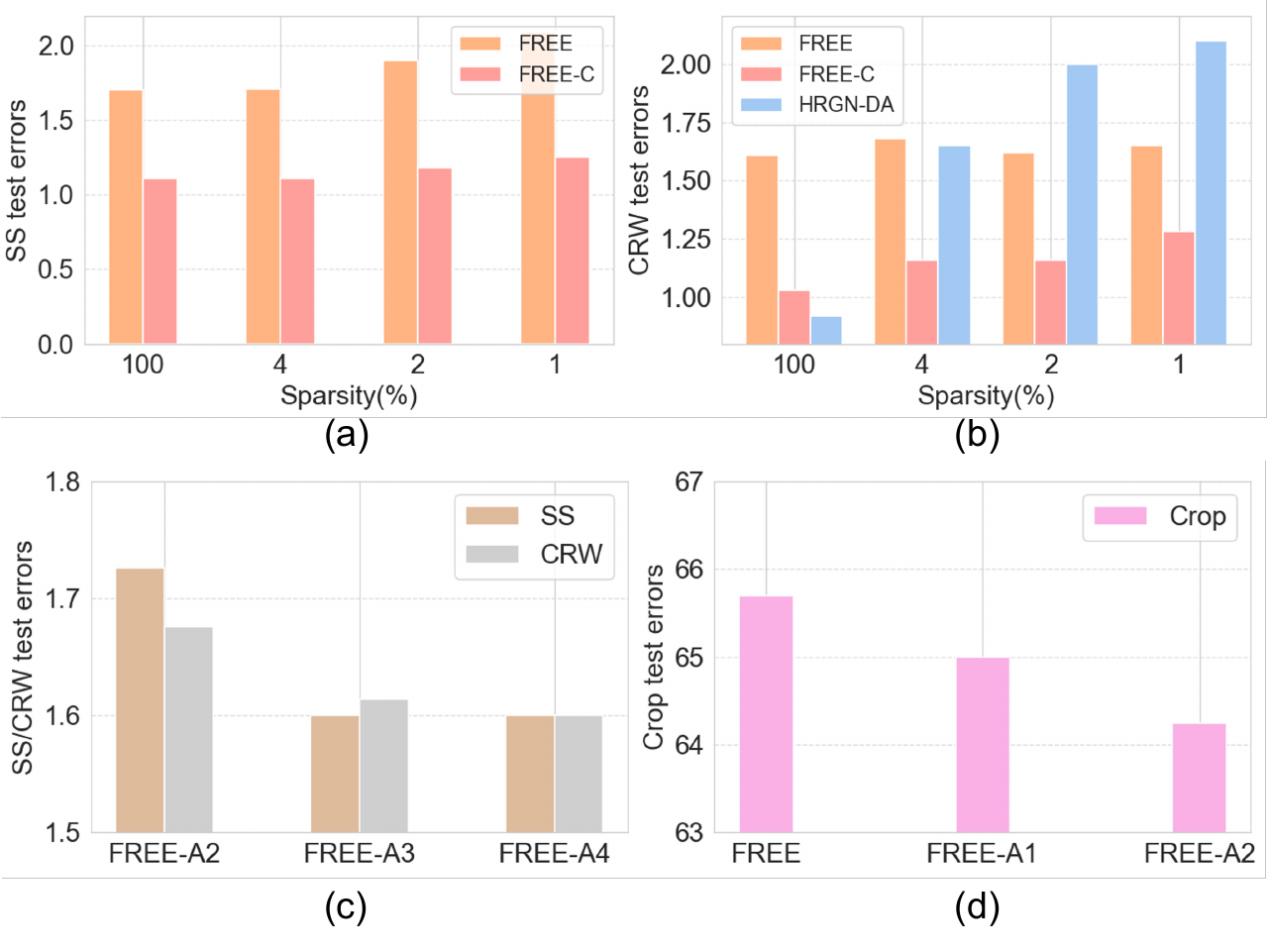}
    \caption{\small Evaluation of FREE with auxiliary information. }
    \label{fig:auxiliary}
\end{wrapfigure}

We evaluate FREE’s ability to harness auxiliary observations and diverse input features. FREE-C refers to the model that incorporates auxiliary observed water temperatures of the current river segment from the prior day (when they are available). We show in Fig.~\ref{fig:auxiliary} (a) and (b) the test errors for predicting stream water temperature upon the inclusion of extra observations under different sparsity levels in SS and CRW, respectively. FREE-C outperforms FREE, demonstrating that our proposed approach can process auxiliary observation data effectively. 
It is worth mentioning that in Fig.~\ref{fig:auxiliary} (b), the baseline HRGN-DA (it is designed for application to a graph of streams, making it incompatible with the SS dataset), which also incorporates new observational data, initially achieves a close performance with FREE-C using the complete training dataset. 
Nonetheless, when the available labeled data is reduced, FREE-C surpasses HRGN-DA in accuracy.
Unlike daily stream temperature prediction, crop yield labels are available on a yearly scale. 
The yield labels from the previous year usually have little impact on the yield of the next year, thus we did not include crop yield experiments here.
On the other hand, we consider variants of FREE using different sets of features.
To mimic different input features in real scenarios, we create data samples with the following assumption. 
All data samples have access to the meteorological features as the daily weather data are publicly available. Different data samples may have different features from other data sources, which are randomly sampled.   
We use FREE-A$m$ to represent the variant of the FREE method that uses the meteorological features and $m$ randomly selected additional features. Fig.~\ref{fig:auxiliary} (c) shows the performance of predicting water temperature upon the integration of additional features, and Fig.~\ref{fig:auxiliary} (d) shows the predictive performance of corn yield upon the integration of additional features. For both datasets, the use of additional features can reduce the prediction errors.

\subsection{Evaluation on model transferability}

\begin{wrapfigure}[16]{r}{0.45\textwidth}
    \centering
    \includegraphics[width=0.45\textwidth]{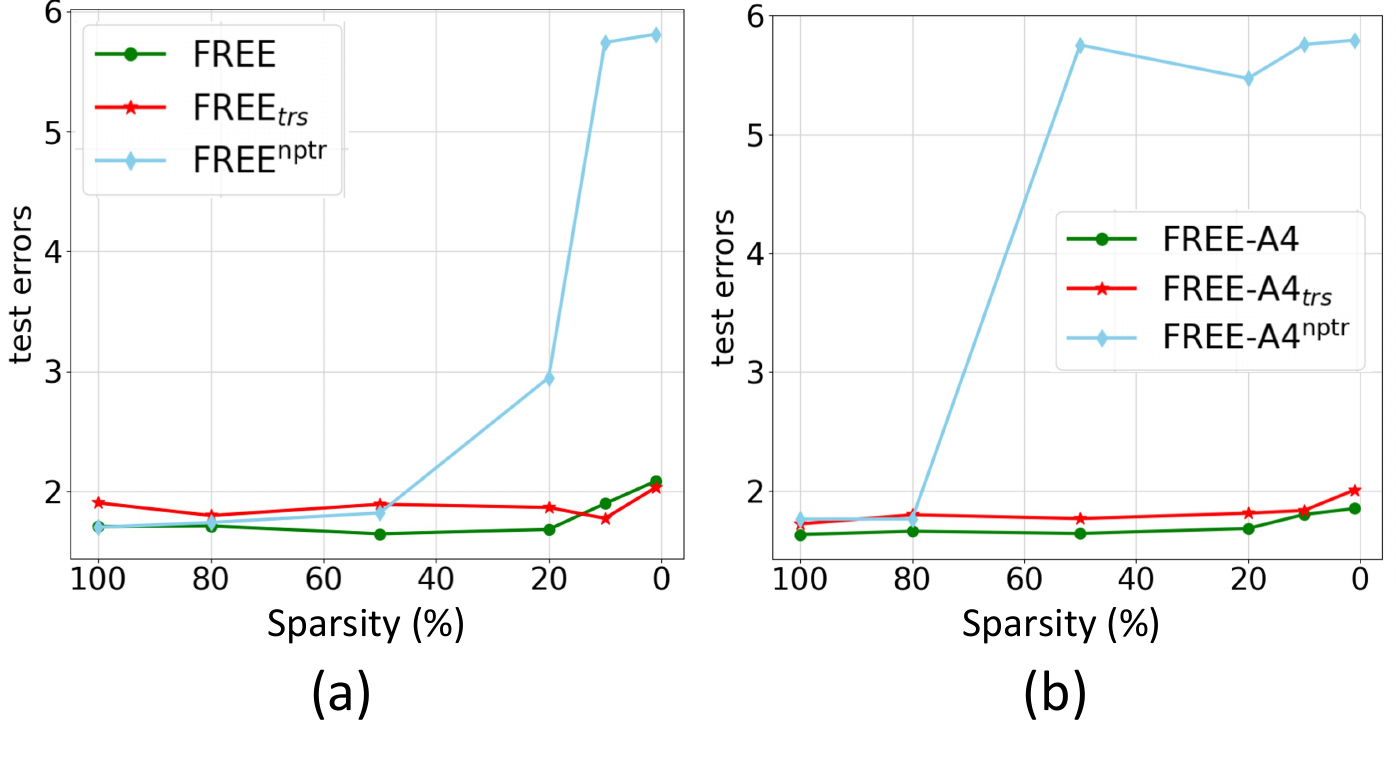}
    \caption{\small Comparison of RMSE under different sparsity levels in the target SS. (a)~train on meteorological features. (b)~trained on meteorological and four additional features.}
    \label{fig:sparsity}
\end{wrapfigure}

We assess transfer learning by adapting CRW-pretrained models to the SS region. 
Specifically, we compare the transferred models pre-trained on CRW (FREE$_{trs}$ and FREE-A4$_{trs}$) with models pre-trained on the same segment in SS (FREE, FREE-A4), and with those trained on observational data directly without pre-training (FREE$^{nprt}$ and FREE-A4$^{nprt}$).  
As Fig.~\ref{fig:sparsity} shows, 
the pre-trained models 
consistently outperform the models without pre-training 
under different data sparsity levels, maintaining stable performance despite reduced label availability. 
Remarkably, the pre-trained model from a different domain can have a similar degree of contribution to performance enhancement compared to the model pre-trained on the same target segment. 
These findings emphasize the 
potential of this approach for building a global pre-trained model over large regions with small training data.  



\section{Conclusion}
This paper introduces FREE, a novel LLM-based framework for environmental modeling that converts different input features and auxiliary observations for different data samples into natural language for semantic recognition. 
FREE is shown to outperform existing methods in the context of predicting stream water temperature and annual crop yield, especially under data sparsity.
The simulation-based pre-training also aids in extracting physically consistent data patterns, which improves the generalizability and transferability to different regions. 
We anticipate the FREE framework to serve as a stepping stone to building foundational solutions for modeling complex environmental and physical systems.  

\section{Acknowledgments}
This work was supported by the National Science Foundation (NSF) under grants 2239175, 2316305, 2147195, 2203581, 2425844, 2425845, 2430978, 2126474, 2530609, and 2530610;
the USGS awards  G21AC10564 and G22AC00266;   the NASA grants 80NSSC24K1061 and 80NSSC25K0013; and the NSF NCAR's Derecho HPC system. This research was also supported in part by the University of Pittsburgh Center for Research Computing through the resources provided. 

\bibliography{reference,Xiaowei}
\bibliographystyle{colm2025_conference}

\end{document}